\documentclass[preprint,12pt,number]{elsarticle}
\usepackage{tikz}
\usetikzlibrary{shapes.geometric, arrows}
\tikzstyle{startstop} = [rectangle, rounded corners, 
minimum width=3cm, 
minimum height=1.3cm,
text centered, 
draw=black, 
fill=red!0]

\tikzstyle{io} = [trapezium, 
trapezium stretches=true, 
trapezium left angle=70, 
trapezium right angle=110, 
minimum width=3cm, 
minimum height=1cm, text centered, 
draw=black, fill=blue!0]

\tikzstyle{process} = [rectangle, 
minimum width=3cm, 
minimum height=1cm, 
text centered, 
text width=3cm, 
draw=black, 
fill=orange!0]

\tikzstyle{decision} = [diamond, 
minimum width=3cm, 
minimum height=1cm, 
text centered, 
draw=black, 
fill=green!0]
\tikzstyle{arrow} = [thick,->,>=stealth]
\usepackage{subfig}
\usepackage{float}
\usepackage{multirow}
\usepackage{multicol}
\usepackage{url}
\usepackage{textcomp}
\usepackage{graphicx}
\usepackage{stackengine}
\def\delequal{\mathrel{\ensurestackMath{\stackon[1pt]{=}{\scriptstyle\Delta}}}}
\usepackage{inputenc}
\usepackage{babel}
\usepackage{amsthm}
\usepackage{amsmath,amsfonts,amssymb}
\usepackage{algorithm}
\usepackage{algpseudocode}
\usepackage{mathtools}
\DeclareMathOperator*{\argmax}{arg\,max}
\DeclareMathOperator*{\argmin}{arg\,min}
\newcommand{\dst}{\displaystyle}
\usepackage{lipsum}
\numberwithin{equation}{section}
\numberwithin{figure}{section}
\numberwithin{table}{section}
\usepackage{enumerate}
\newtheorem{theorem}{Theorem}[section]

\newtheorem{lemma}[theorem]{Lemma}
\newtheorem{claim}[theorem]{Claim}

\begin{document}

	

	\begin{frontmatter}
		\title{Optimal Projections for Discriminative Dictionary Learning using the JL-Lemma\footnote{Preprint submitted to arxiv \cite{arxiv_JLpaper}.}}
		
		\author[inst1]{G.Madhuri}
		\ead{17mcpc05@uohyd.ac.in}
		\author[inst1]{Atul Negi}
		\ead{atul.negi@uohyd.ac.in}
		\author[inst2]{Kaluri V. Rangarao}
		\ead{kaluri@ieee.org}
		
		\affiliation[inst1]{
			organization={School of Computer and Information Sciences},
			addressline={University of Hyderabad}, 
			city={Hyderabad},
			postcode={ 500046},
			state={Telangana},
			country={India}.
		}
		\affiliation[inst2]{organization={Visiting Scientist, School of Computer and Information Sciences},
			addressline={University of Hyderabad}, 
			city={Hyderabad},
			postcode={ 500046},
			state={Telangana},
			country={India}.
		}
		\begin{abstract}
			Dimensionality reduction-based dictionary learning methods in the literature have often used iterative random projections. The dimensionality of such a random projection matrix is a random number that might not lead to a separable subspace structure in the transformed space. The convergence of such methods highly depends on the initial seed values used. Also, gradient descent-based updates might result in local minima. This paper proposes a constructive approach to derandomize the projection matrix using the Johnson-Lindenstrauss lemma. Rather than reducing dimensionality via random projections, a projection matrix derived from the proposed Modified Supervised PC analysis is used. A heuristic is proposed to decide the data perturbation levels and the dictionary atom's corresponding suitable description length. The projection matrix is derived in a single step, provides maximum feature-label consistency of the transformed space, and preserves the geometry of the original data. The projection matrix thus constructed is proved to be a JL-embedding. Despite confusing classes in the OCR datasets, the dictionary trained in the transformed space generates discriminative sparse coefficients with reduced complexity. Empirical study demonstrates that the proposed method performs well even when the number of classes and dimensionality increase. Experimentation on OCR and face recognition datasets shows better classification performance than other algorithms. 
		\end{abstract}
		
	\end{frontmatter}
	
	\section{Introduction}
	The Sparse Representation (SR) model finds the sparse latent feature space, avoiding hand-crafted feature extraction. The SR model is similar to the receptive field properties of the visual cortex in mammals \cite{olshausenfield}. The use of overcomplete ($\#columns > \#rows$) learned dictionaries \cite{ksvd, Mairal_supervisedDL} in the SR model became popular for richer representations and classification. For high-dimensional signal classification, discriminative sparse feature extraction requires training a shared global dictionary. 
	To lower the complexity of training the dictionary for classifying high dimensional data from many classes, constraints like low-rank, sparsity are imposed on the coefficient matrix. Though compact, the trained dictionary must generate discriminative coefficients for classification. Low dimensional atoms (columns) with global and local features of all the classes result in a compact shared dictionary, saving space and time complexity.
	 
	PCA for dimensionality reduction embeds data into a lower dimensional space with a random choice of the number of principal components. The principal components are derived from the data. On the contrary, the JL-lemma prescribes the ambient dimensionality to preserve the pairwise distances between datapoints, but does not depend on the features of the dataset. The proposed JLSPCADL method complements both the methods by deriving the JL-prescribed number of principal components from Modified Supervised PCA (MSPCA). 

	%
In the literature, dimensionality reduction-based DL methods apply iterative random projections with a random number of principal components for mapping to a lower dimensional space \cite{SDRDL, JDDRDL}. Instead of using random projections for dimensionality reduction, this article proposes a mathematically sound constructive approach to design a derandomized projection using the Johnson-Lindenstrauss (JL) lemma-prescribed dimensions. The JL-lemma \cite{JLlemma} gives the minimal number of dimensions $p$ required to transform data into a lower dimensional space using Gaussian random projections without exceeding a bounded perturbation of the original geometry. However, such random projections do not guarantee feature label consistency.   In this context, we propose a heuristic to determine the suitable data perturbation threshold and the corresponding dimensionality from the JL-lemma \cite{JLlemma_proof2003} as a novel practical utility. The JL-lemma prescribed dimensionality $p$ becomes the Suitable Description Length (SDL) of dictionary atoms in the transformed space. 
	
	If two data points are far from each other in the original space but very close in the transformed space, it is highly likely that these two are coded using the same set of dictionary atoms, leading to misrepresentation and misclassification. To avoid such embeddings, we also propose Modified-SPCA (MSPCA) to derive a constructive projection matrix with $p$ orthonormal principal components.  This transformation is designed to maximize the dependence between the data and the labels, based on \textit{Hilbert-Schmidt Independence Criterion (HSIC)} \cite{supPCAPR2011} for Reproducing kernel Hilbert Spaces (RKHS). 
	Unlike Supervised PCA \cite{supPCAPR2011} where the number of principal components is randomly chosen, the proposed Modified Supervised PCA (M-SPCA) with the suitable number of principal components ($p$) gives the optimal transformation matrix for each dataset. 
	Learning a dictionary with atoms of dimensionality $p$ (SDL) in the transformed space results in atoms with local and global features, and the corresponding discriminative coefficients are used as features for better classification performance. Subspace RIP guarantees the separation of classes in the projected space \cite{RIPwithJLIEEESP2020}. An argument supporting the Subspace Restricted Isometry Property (RIP) in the  M-SPCA-embedded space and a detailed proof of the same is given in Section \ref{proposedsection}. 
	\subsection{Highlights of JLSPCADL}
	\begin{itemize} 
		\item A heuristic to determine the optimal data perturbation threshold and the corresponding interval for optimal dimensionality $p$ based on the JL-lemma is proposed. This optimal dimensionality is the SDL for dictionary atoms.
		\item A constructive approach is given to obtain the optimal derandomized transformation matrix when $p \leq d$ and when $p>d$ using Modified Supervised PCA. 
		\item Maximum feature-label consistency is achieved with the proposed projection matrix where the $p$ principal components include label information.
		\item Mathematically proved that the proposed transformation matrix is a JL-embedding and satisfies subspace RIP, i.e., the pairwise distances between subspaces are preserved in the transformed space.
		
	\end{itemize} 
	The notation used in this article is given in Table \ref{notation}.
	\begin{table}[h]
		\footnotesize
		\centering
		\caption{Notation.} \label{notation}  
		\begin{tabular}{c c}
			
			\hline
			
			Matrix & $A-Z$ \\
			Vector & $\bar{a}-\bar{z}$\\
			$\|.\|_p$ or $l_p$-norm, $\|\bar{a}\|_p=$ & $(\sum\limits_{j=1}^n |a_{j}|^p)^{(1/p)}$\\
			Frobenius norm, $\|A\|_F=$& $\sqrt{\sum\limits_{i=1}^m \sum\limits_{j=1}^n |a_{ij}|^2}$ \\
			Inverse and Transpose & Superscripts $-1$ and $T$ \\
			Identity matrix&$I$ \\
			Pdf of a Gaussian multivariate & $\mathcal{N}(X; \bar{\mu}, \Sigma)$ \\ 
			Trace of matrix $A$ & $tr(A)$\\ 
			No. of samples/class &  $\#S_C$\\
			No. of iterations & $n$\\
			Projection dimension & $p$\\
			\hline
		\end{tabular} 
	\end{table}
	\subsection{Sparse Representation Problem} 
	Given a set of $N$ training images $Y=\{\bar{y}_i,i=1,\ldots,N|$ $\bar{y}_i \in \mathbb{R}^{d\times 1}\}$ from $C$ classes, and the corresponding label matrix $H\in \mathbb{R}^{C\times N}$, the problem is to find a dictionary $D\in \mathbb{R}^{d\times K}$ and the corresponding coefficient matrix $X\in \mathbb{R}^{K\times N}$ such that $Y\approx DX + \epsilon$. Using $l_1-$norm regularization, the SR optimization problem is 
	\begin{equation}\label{SRobjfn}
	<\hat{D},\hat{X}>= \argmin_{D,X}(\|Y-DX\|_F^2 + \lambda \|X\|_1 ).
	\end{equation} subject to $\|d_j\| \leq 1, \forall j=1,2,\ldots,K$. Here $\hat{D},\hat{X}$ denote the optimal values of $D,X$. The unit norm constraint on dictionary columns avoids arbitrarily large coefficients in $X$. The optimization problem in equation \eqref{SRobjfn} can be solved by updating coefficient matrix X w.r.t a fixed dictionary $D$ (called \textit{Sparse Coding}) and updating $D$ w.r.t a fixed $X$ (called \textit{Dictionary Learning}, DL). Here, the objective is to learn a discriminative dictionary using the training samples and their labels with maximum feature-label consistency. When the number of classes is high, learning a shared discriminative dictionary along with a classifier \cite{dksvd},\cite{LCKSVD},\cite{Yang2011FisherDD} becomes computationally viable. Sparse coefficients learned w.r.t. a shared discriminative dictionary are powerful discriminative features for classification as demonstrated in \cite{SCMLP2019}. However, the dictionary parameters, i.e., the description length (dimensionality) of atoms and the number of atoms, must be optimal for reduced complexity and improved performance.

	We discuss related work in section \ref{relatedwork} and the JL-lemma application in section \ref{JL-lemmasection}, the proposed method for dictionary learning, and a detailed proof that the proposed projection matrix transforms data such that the Subspace RIP holds is discussed in Section \ref{proposedsection}. The classification rule is explained in section \ref{sec:propclassifyrule}, and experimental results are presented in section \ref{expresults}. Discussion and conclusions are presented in Section \ref{discuss} and Section \ref{conclusions}, respectively.  Modified KSPCA (M-KSPCA) and its Complexity is explained in the Appendix.
	
	\section{DL in reduced dimensionality space: Related work}\label{relatedwork}
	There have been attempts to first reduce the dimensionality of data and then learn the dictionary in the transformed subspace \cite{DLinoptimalmetricspace}. It is important to find the correct embedding or projection of data so that the dictionary learned in the transformed space generates useful coefficients as data features. In Kernelized Supervised DL (KSDL) \cite{kernelizedsupDL}, Supervised PCA (SPCA) has been applied to the training data to get an orthonormal transformation matrix whose transpose is used as the dictionary. Here, the description length of the atoms is reduced to a random number. In \cite{SDPDLPR}, a gradient descent-based approach is given for discriminative orthonormal projection of original data into a subspace and simultaneous dictionary learning for Sparse Representation based Classification (SRC \cite{Face_rec_via_sparserep}). However, the method becomes computationally intensive when there are many classes and might result in suboptimal solutions. JDDRDL \cite{JDDRDL} jointly learns the projection matrix for reducing dimensionality and a discriminative dictionary iteratively. SDRDL \cite{SDRDL} simultaneously learns the projection matrix for each class and the corresponding class-wise dictionaries iteratively. As the number of classes increases, generating a sparse coefficient of test signal w.r.t each class-specific dictionary consumes time. The above methods try to reduce the dimensionality using an initialization of the projection operator and simultaneously learn the dictionary and coefficients iteratively until convergence. If this initial guess is far from the optimal, convergence is delayed due to accumulated error. A sparse embedded dictionary is learned in the embedded space by imposing a Fisher discriminant-like criterion using a sparse coefficient similarity matrix based on their labels in \cite{SparseembedDLFace2017PR}. 
	The similarity matrix is given as \[S_{ij}= 1/N_{c} \text{ when } label(x_i)=label(x_j)=c,\] else \[S_{ij}= 1/(N-N_{v}) \text{ where } v=label(x_i).\] Instead of optimizing the projection, the coefficients are optimized to decrease inter-class similarities.
	
	In this work, we propose to use the JL-lemma to determine the number of features required so that the original clusters are preserved and use this optimal dimensionality for the projection of a dataset as the Suitable Description Length (SDL) of dictionary atoms. The columns of the projection matrix are derived using Modified-SPCA, which uses SDL as the number of principal components. We prove that this projection matrix is a JL-embedding and preserves distances between subspaces in the transformed space.
	
	\section{The Johnson-Lindenstrauss Lemma}\label{JL-lemmasection}
	Reducing the dimensionality before learning the dictionary enables computing with limited resources and avoids model over-fit. Each $p$ of a dataset depends on the dataset size $N$ and the required data perturbation threshold $\epsilon$. Depending on the choice of data perturbation and the dataset size, JL-lemma provides the optimal dimensionality $p$ to project the dataset.
	\begin{lemma}{JL-Lemma \cite{JLlemma_proof2003}:}
		Given a set of $N$ data points in $\mathbb{R}^d$ and $0< \epsilon <1$, if $\dst p\geq \frac{12\log N}{\epsilon^2(1.5-\epsilon)}$, then there exists a map $\dst f:\mathbb{R}^d\to \mathbb{R}^p$ such that 
		\begin{eqnarray*}\label{JLlemma}
			\|f(x_i)-f(x_j)\|_2^2 &\leq &(1-\epsilon,1+\epsilon)\|x_i-x_j\|_2^2,\\
			1 &\leq &i,j \leq N.
		\end{eqnarray*}
	\end{lemma}
	A proof of JL-lemma in \cite{JLlemma_proof2003} considers $f$ to be a random matrix with entries independently identically drawn from a Gaussian distribution with mean 0 and variance 1. Such a map is guaranteed to preserve distances in the transformed space up to a scale factor of distances between original pairs with probability at least $\frac{1}{N}$ \cite{JLlemma_proof2003}. 
	In \cite{RIPwithGaussianrandomprojTSP2018}, the authors theoretically prove the RIP of \textit{Gaussian random projections}  when the dimension after projection is based on the JL-lemma. In \cite{derandomizeJLlemma}, a Gaussian random projection matrix $\dst U:\mathbb{R}^d \to \mathbb{R}^p$ maps any vector $\Bar{y} \in \mathbb{R}^d$ to $\dst \frac{U^T \Bar{y}}{\sqrt{p}}$ with a perturbation of atmost $\dst (1\pm \epsilon)$, while projecting the vectors into $\dst \frac{2}{\epsilon^2}\log 2N$ dimensions. 
	\subsection{Derandomization of the Projection matrix}
	Gaussian random projections using the JL-lemma satisfy the Restricted Isometry Property \cite{RIP_JL_2020}. The methods proposed to derandomize the projection matrices in \cite{achlioptas2001database, ACHLIOPTAS_JL_binarycoin2003} try to reduce the number of random bits used to project the data by proposing simple probability distributions to fill the entries of the projection matrix. Though the conditions of orthogonality and normality are not imposed on the columns of the transformation matrix, orthonormality is nearly achieved. In \cite{Arriagavempala}, the JL-projection matrix entries are from $\{\pm 1\}$, without any bound on $p$.  In \cite{JLlemma_proof2003}, Gaussian random projection projecting to a dimensionality of $p \geq \frac{24 \log N}{3\epsilon^2 -2\epsilon^3}$ is proved to be a JL-embedding. In \cite{FastJLT2006}, a product of two random and a deterministic Hadamard matrix gives Fast JL-transform. These projections are not data-dependent and do not guarantee feature-label consistency in a supervised scenario. Here, we propose to use the lower bound on the JL-embedding dimensionality given in \cite{JLlemma_proof2003} in the map given in \cite{derandomizeJLlemma} i.e. $\Bar{y} \to U^T\Bar{y}/\sqrt{p}$  where $U$ is derived from M-SPCA. 
	
	\section{Proposed method: JLSPCADL}\label{proposedsection} 
	The proposed method starts with calculating the optimal data perturbation $\epsilon$ and the optimal projection dimension $p$. Then the transformation matrix with $p$ principal components is obtained from M-SPCA ($p\leq d$) or M-KSPCA ($p>d$).
	The shared dictionary and the corresponding discriminative sparse coefficients are learned in the transformed space. The framework of JLSPCADL is given in  Fig. \ref{frameworkfigure}.
	\begin{figure}
		\begin{tikzpicture}[node distance=2cm,thick,scale=0.75, every node/.style={scale=.85}]
		\node (start) [startstop] {Start};
		\node (in1) [io, below of=start] {Input: $Y_{d \times N}$, Labels $H\in \{0,1\}^{C \times N}$};
		\node (JL-) [process, below of=in1]{$p\geq \frac{24 \log N}{3\epsilon^2-2\epsilon^3}$ where $\epsilon \in [0.3,0.4]$};
		\node (pro1) [process, right of=in1, xshift=3.5cm] {Find $U_{d \times p}$ using M-SPCA};
		\node (out1) [io, above of=pro1] {$U$};
		\node (trans)[process, right of=out1, xshift=1.5cm]{$U^TY=Z_{p \times N}$};
		\node (pro2) [process, right of=pro1, xshift=1.5cm] {Learn $D,X$ using K-SVD, M-SBL such that $Z \approx DX$};
		\node (out2) [io, below of=pro2] {$D,X$};
		\node (pro3) [process, below of=out2] {Compute medoids of $X_{K\times N}$ as cluster centers.};
		\node (out3) [io, below of=pro3] {Label(test-image) using classification rule};
		\node (stop) [startstop, below of=out3] {Stop};
		
		\draw [arrow] (start) -- (in1);
		\draw [arrow] (in1) -- (JL-);
		\draw [arrow] (JL-) -- (pro1);
		\draw [arrow] (in1) -- (pro1);
		\draw [arrow] (pro1) -- (out1);
		\draw [arrow] (out1) -- (trans);
		\draw [arrow] (trans) -- (pro2);
		\draw[arrow](pro2) -- (out2);
		\draw [arrow] (out2) -- (pro3);
		\draw [arrow] (pro3) -- (out3);
		\draw [arrow] (out3) -- (stop);
		\end{tikzpicture}
		\caption{Framework of proposed JLSPCADL for classification: $p$ is determined from $N$ and $\epsilon \in [0.3,0.4]$, $U$ from M-SPCA, $D,X$ using K-SVD in the transformed space $Z$, and finally the classification label using \eqref{eq:classifyrule}.}\label{frameworkfigure}
		\vspace{-0.5cm}
	\end{figure}
	\subsection{Determination of optimal $\epsilon $ and $p$} \label{graphicalp_eps}
	Let $\dst p: \epsilon \in (0,1) \to  \frac{12\log N}{\epsilon^2(1.5-\epsilon)}+1,$ with $\epsilon=0$ being the singularity of $p$. $ \dst\frac{dp}{d\epsilon} = \frac{36 \log N(\epsilon -1)}{\epsilon^3(1.5-\epsilon)^2}$. 
	
\begin{figure}
		\centering
		\vspace*{-2mm}
		
		\subfloat[\footnotesize If $\epsilon \to 0$, number of dimensions increase]{\includegraphics[height=3cm,width=4.1cm]{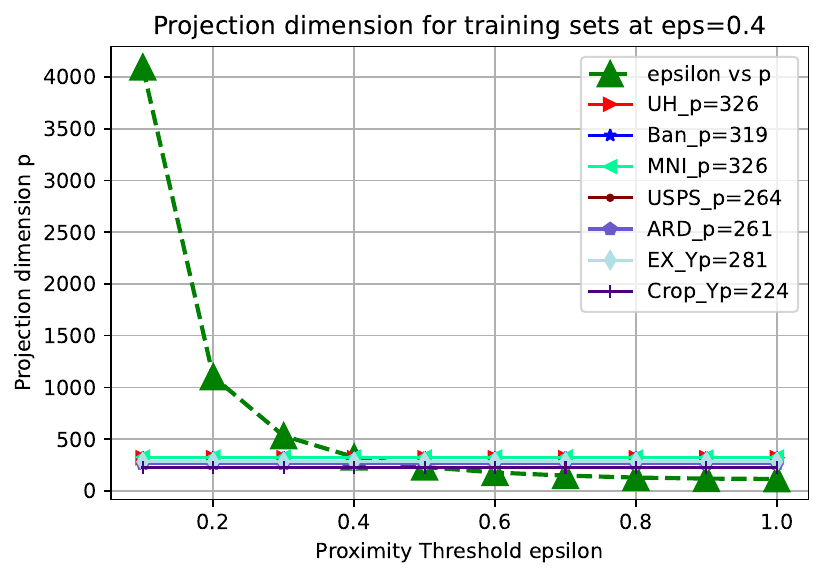}}
		\subfloat[\footnotesize As $p$ does not change much when $\epsilon > 0.4$, ideally $\epsilon \in {[0.3,0.4]}$. ]{\includegraphics[height=3cm,width=4.2cm]{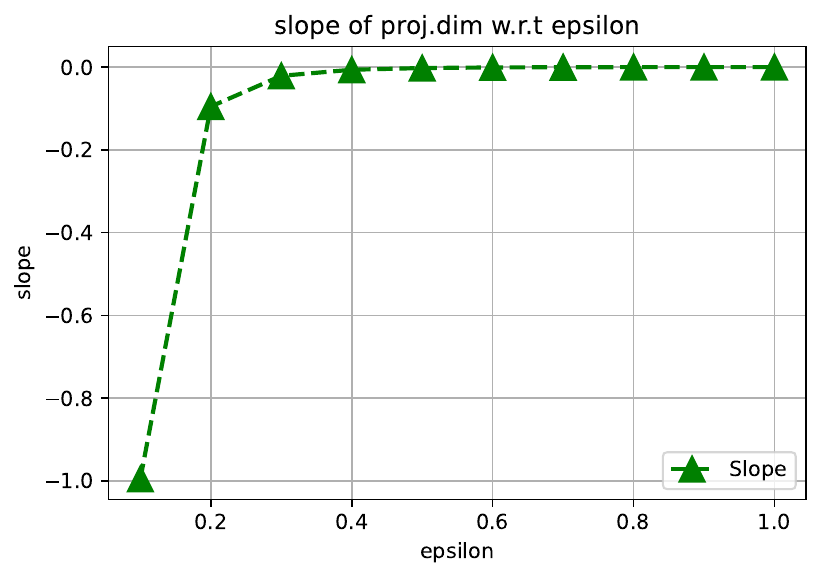}}
		\subfloat[\footnotesize When $\epsilon=0.9$, distances between ARDIS points blowup after applying PCA with p=92. ]{\includegraphics[height=3cm,width=4.2cm]{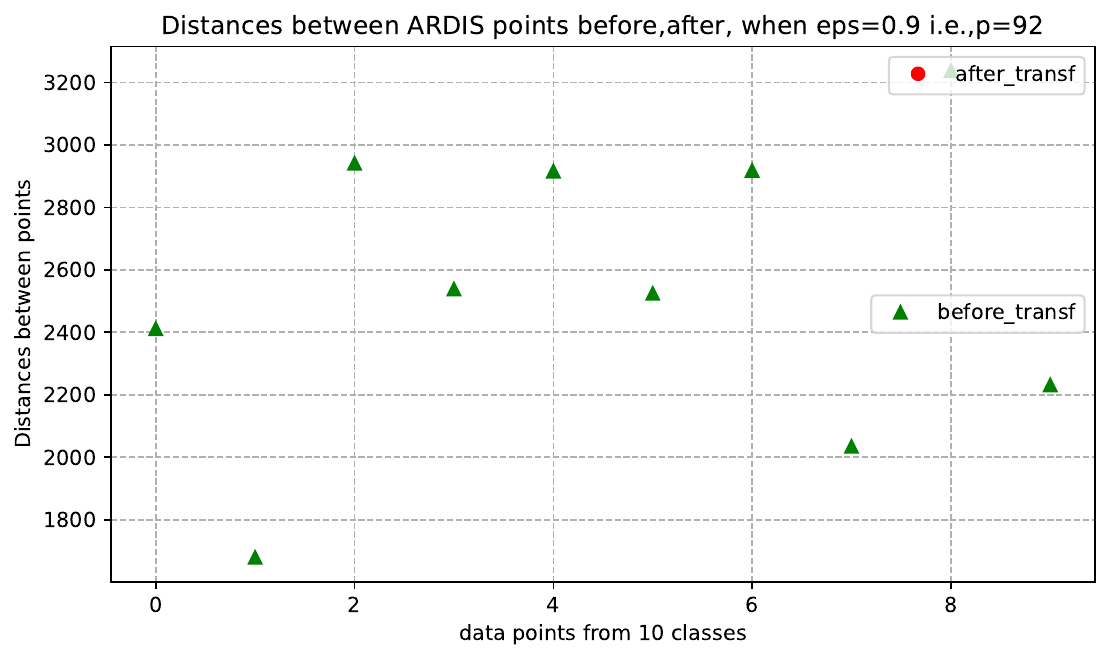}
			
		}
		
		\caption{(a)Lower bounds on $p$ for $\epsilon=0.4$ when the curve flattens. (b)$\frac{dp}{d\epsilon}$ vs $\epsilon$. The projection dimension of datasets is chosen at the point where the curve in (b) starts to flatten (c) If $\epsilon$ is closer to 1, then the distance between the mapped data points (solid red line ), $\|f(x_i)-f(x_j)\|_2^2, \forall i,j $, could blow up. }\label{fig:pversusepsilon}
		\vspace{-0.5cm}
	\end{figure}

	If $\epsilon$ is closer to 0, then the value $p$ increases, which is undesirable. If $\epsilon$ is closer to 1, then the distance between the mapped data points (solid red line ), $\|f(x_i)-f(x_j)\|_2^2, \forall i,j $, could blow up as shown in Fig.\ref{fig:pversusepsilon} (c). $\frac{dp}{d\epsilon} \to 0$ as $\epsilon \to 1$ which is outside the domain. Instead of arbitrarily selecting $\epsilon$, we look at the point where $\frac{dp}{d\epsilon} \to 0$. Based on the data size, the optimal perturbation threshold $\epsilon$ used in JL-lemma is selected as the point where the value of $\frac{dp}{d\epsilon}$ tends to zero. This indicates that the value of $p$ does not change much after this point.
	Thus, the optimal perturbation threshold interval considered here is $[0.3,0.4]$, and the corresponding $p$ is the optimal projection dimensionality interval. From our experiments, it is found that the value of $p$ beyond this interval gives sub-optimal results either in terms of classification accuracy, computing time, or both. 
	Table \ref{derivedparams} presents how different parameters of DL are determined mathematically and statistically. 
	\begin{table}
		\footnotesize
		\caption{Mathematical Determination of Parameters.} \label{derivedparams}   
		\begin{center}
			\begin{tabular}{c|c}
				\hline
				Parameter & Determined by\\
				\hline
				$p$ & Using JL-lemma \\
				$\epsilon$ & Graph of $\epsilon$ vs $\frac{dp}{d\epsilon}$\\
				$U$ & Using Modified SPCA\\
				$K$& $\#S_C \times \#Classes$\\
				Sparsity & Automatic relevance determination using Multiple snapshot-SBL \cite{SBL_Tipping} \\
				\hline
			\end{tabular}
		\end{center}   
	\end{table}
	
	For example, the UHTelPCC dataset has $N=50000$ samples for training. We consider the value of $p$ corresponding to $\epsilon \in [0.3,0.4]$, decreasing from 522 to 320. This implies that $p \geq 320$ for $\epsilon=0.4$, preserves the clusters such that
	\[\|f(x_i)-f(x_j)\|_2^2 \in (0.6, 1.4)\|x_i - x_j\|_2^2\].
	For each dataset, we use $p$ as the number of principal components in Modified-SPCA to transform data.  Fig. \ref{fig:pversusepsilon}(a) gives lower bounds of projection dimensions of the datasets used for experimentation.
	Fig. \ref{fig:pversusepsilon} (c) shows pairs of data points from ten classes on the horizontal axis and the distances between them on the vertical axis.
	\subsection{Supervised PCA}
	Supervised PCA \cite{supPCAPR2011} gives a transformation matrix $U$ based on HSIC criterion. The objective is to find 
	\begin{equation}\label{closedformsol}
	\Hat{U} = \argmax_U tr(U^TYCLCY^TU)
	\end{equation} sub. to $U^TU=I$, where $L=HH^T$ and $C$ is the data centering matrix. This problem has a closed-form solution from known linear algebra Lemma \ref{lemma1}.
	\begin{lemma}\label{lemma1}
		If $\Hat{U}= \argmax\limits_U{tr(U^TQU)}$, then the columns of $U$ are the eigenvectors corresponding to the largest p eigenvalues of $Q$.
	\end{lemma}
	%
	\subsection{Transformation of data using Modified-SPCA (M-SPCA)}
	
	We propose Modified-SPCA, which finds the projection matrix with the number of principal components $p$ determined using the JL-lemma as explained in section \ref{graphicalp_eps}. By scaling the data $Y$ to standard mean and variance, we avoid multiplying with the centering matrix C and thus reduce the complexity. The transformed space $U^TY$ has maximum dependence on the label matrix $H$. Therefore, from HSIC criterion \cite{supPCAPR2011}, the optimal $U$ is
	\begin{equation}\label{UequnwoC}
	\begin{split}
	\hat{U}&= \argmax_U \{tr((U^TY)^T(U^TY)H^TH)\}\\
	&= \argmax_U tr(\underbrace{Y^TU}\underbrace{U^TYL}) \\
	& =\argmax_U tr(U^T\underbrace{YLY^T}U).
	\end{split}
	\end{equation} 
	sub. to $U^TU=I$ where $L=H^TH$ is the label kernel matrix.
	Using Lemma \ref{lemma1}, we obtain $U$ whose columns are $p$ eigenvectors corresponding to the $p$ largest eigenvalues of $YLY^T$. $YLY^T$, being a real symmetric matrix, has real eigenvalues, and the corresponding eigenvectors are orthogonal to each other. Constituting $U_{d\times p}, p\leq d $ with a set of orthonormal eigenvectors from these eigenvectors, $U$ is a semi-orthogonal matrix with unit norm columns such that $U^TU=I$.
	The objective of \textbf{JLSPCADL} is to find
	\begin{eqnarray}\label{objectivefn}
	<\hat{U},\hat{D},\hat{X}>& = &\argmin_{U,D,X}\big\{\|U^TY-DX\|_F^2 \nonumber \\
	& + &\lambda g(X) -tr(U^TYLY^TU) \nonumber \\&+ &\|U^TU-I\|_F^2\big\} 
	\end{eqnarray}
	such that $\|d_j\|_2^2=1 \forall j = 1,2,\ldots, K$. Here $\lambda$ is a regularization parameter.\\
	The first term in equation \eqref{objectivefn} is the data fidelity term in the transformed space. The second term is a constraint on the coefficients, here sparsity. The third and fourth terms are independent of $D$ and $X$, and hence, $U$ can be deduced in a single step using M-SPCA, with the number of principal components $p$ derived from the JL-lemma above. The transformation $U$ contains $p$ PCs of $YLY^T$, orthogonal to each other. Thus, $U$ is a semi-orthogonal transformation that preserves distances and angles between data points. However, a detailed proof that $U$ is a JL-embedding is given here
	
	From \cite{RIPwithJLIEEESP2020}, we have the following results on random projection matrices.
	\begin{theorem}\label{JLproperty}
		Johnson-Lindenstrauss (JL) Property: A random matrix $A\in \mathbb{R}^{p\times d}$ is said to satisfy JL property if there exists some positive constant $\Tilde{c}$ such that for any $0<\epsilon<1$ and for any $\Bar{x} \in \mathbb{R}^d$, $\dst P(|\|A\Bar{x}\|_2^2-\|\Bar{x\|_2^2}| >\epsilon\|\Bar{x}\|_2^2) \leq 2 e^{-\Tilde{c}\epsilon^2p}$. 
	\end{theorem}
	\begin{lemma}\label{subspaceRIPlemma}
		Assuming the random matrix $A$ satisfies the JL property, the pairwise distances between the subspaces in the projected space are preserved.    
	\end{lemma}
	By proving that the proposed derandomized projection, $U$, satisfies JL-property in Theorem \ref{JLproperty}, the pairwise distances between the subspaces in the projected space are preserved as stated in Lemma \ref{subspaceRIPlemma}.
	\begin{lemma}\label{proofofJLproperty}
		The proposed projection matrix $U_{d\times p}=[\hat{e}_i]_{i=1}^{p}$ contains $p$ orthonormal eigenvectors of $YLY^T$ and any $\bar{y} \in \mathbb{R}^d $ is mapped to $\frac{U^T\Bar{y}}{\sqrt{p}}$ i.e. $\Bar{z}=\frac{U^T\Bar{y}}{\sqrt{p}}$. Then, the following conditions hold. 
		\\
		(a) $\dst E[\|\Bar{z}\|_2^2]\leq E[\|\Bar{y}\|_2^2]$\\
		(b) $\dst P(\|\Bar{z}\|_2^2 \notin (1\pm \epsilon)\|\Bar{y}\|_2^2) \leq 2e^{-\Tilde{c}\epsilon^2p}$. 
		Thus, Subspace RIP holds as $U$ satisfies JL property (Theorem $\ref{JLproperty}$).
		\begin{proof} (a) Using the Cauchy-Schwarz inequality, for any $\Bar{y} \neq \Bar{0} \in \mathbb{R}^d$, we have
			\begin{equation}\label{eq:matrix_l2norm}
			\begin{split}
			\frac{\|U^T\Bar{y}\|_2^2}{\|\Bar{y}\|_2^2} & \leq 
			\|U^T\|_2^2 
			\end{split}
			\end{equation}
			
			\begin{equation}
			\begin{split}
			E[\|\Bar{z}\|_2^2] &= \frac{1}{p}E[\|U^T\Bar{y}\|_2^2]\leq \frac{1}{p}E[\|U^T\|_2^2 \|\Bar{y}\|_2^2] \leq \frac{1}{p}E[\|U^T\|_F^2 \|\Bar{y}\|_2^2]\\
			& = \frac{1}{p}E[trace(I_p)\|\Bar{y}\|_2^2] \\ 
			&=E[\|\Bar{y}\|_2^2]
			\end{split}
			\end{equation}
			(b) Using $\log(1-x)=-x-\frac{x^2}{2}-\frac{x^3}{3}- \ldots$ and
			\begin{equation}
			\begin{split}
			\|U\|_2 
			\delequal \sup_{\Bar{w} \neq \Bar{0}} \frac{\|U^T\Bar{w}\|_2^2}{\|\Bar{w}\|_2^2}=\lambda_{max}
			\geq 
			\frac{\|U^T\Bar{y}\|_2}{\|\Bar{y}\|_2}
			\end{split} 
			\end{equation}
			For any $\Bar{y} \neq \Bar{0} \in \mathbb{R}^d$, we have
			\begin{equation}
			\begin{split}
			P(\|\Bar{z}\|_2^2 \leq (1-\epsilon)\|\Bar{y}\|_2^2 )= P(\frac{\|U^T\Bar{y}\|^2}{p} \leq (1-\epsilon)\|\Bar{y}\|^2)\\
			=P(e^{-\lambda \frac{\|U^T\Bar{y}\|_2^2}{\|\Bar{y}\|_2^2}} \geq e^{-\lambda(1-\epsilon)p})\\
			\leq \frac{E[e^{-\lambda \frac{\sum_{j=1}^{p}|z_j|_2^2}{\|\Bar{y}\|_2^2}}]}{e^{-\lambda (1-\epsilon)p}}
			\end{split}
			\end{equation} 
			Let $t_j= \frac{|z_j|}{\|\Bar{y}\|}$.  $\forall \Bar{y} \in \mathbb{R}^d, \frac{\|U^T\Bar{y}\|}{\|\Bar{y}\|} \leq \|U\|_2= \lambda_{max}$.
			According to the Central Limit Theorem, $t_j \sim \mathcal{N}(\mu_j, \sigma_j), j=1,2,\ldots,p$. Let $t_{1j}= \frac{t_j-\mu_j}{\sigma_j} \sim \mathcal{N}(0,1), j=1,2,\ldots,p$.
			\begin{equation}
			E[e^{-\lambda t_{1j}^2}]= \int_{-\infty}^{+\infty}e^{-\lambda t_{1j}^2}\frac{e^{\frac{-t_{1j}^2}{2}}}{\sqrt{2\pi}}dt_{1j}
			\end{equation}
			Therefore,
			\begin{equation}
			P(\|\Bar{z}\|_2^2 \leq (1-\epsilon)\|\Bar{y}\|_2^2 ) \leq \frac{\prod_{j=1}^{p}E[e^{-\lambda t_{1j}^2}]}{e^{-\lambda (1-\epsilon)p}}\\
			= \frac{1}{(\sqrt{1+2\lambda})^pe^{-\lambda (1-\epsilon)p}}
			\end{equation}

			Choose $\lambda=\frac{\epsilon}{2(1-\epsilon)}>0$ for $0<\epsilon<1$.Therefore,\\
			\begin{equation}
			\begin{split}
			P(\|\Bar{z}\|_2^2 \leq (1-\epsilon)\|\Bar{y}\|_2^2 )&= \frac{(\sqrt{1-\epsilon})^p}{e^{-\epsilon p/2}}= ((1-\epsilon)e^{\epsilon})^{\frac{p}{2}}\\
			&< (e^{-\epsilon-\frac{\epsilon^2}{2}}e^{\epsilon})^{\frac{p}{2}}= e^{-\epsilon^2p/4}\\
			&= e^{-\Tilde{c}\epsilon^2p},
			\end{split}
			\end{equation}\\ where $\Tilde{c}=\frac{1}{4}>0$.\\
			
			Similarly, choose $\dst 0<\lambda=\frac{\epsilon}{2(1+\epsilon)}<\frac{1}{2}$ to get\\
			\begin{equation}
			\begin{split}
			P(\|\Bar{z}\|_2^2 \geq (1+\epsilon)\|\Bar{y}\|_2^2) & \leq 
			\frac{(1+\epsilon)^{\frac{p}{2}}}{e^{\frac{p\epsilon}{2}}}< \frac{(e^{(\epsilon - \frac{\epsilon^2}{2})})^{\frac{p}{2}}}{e^\frac{p\epsilon}{2}}\\
			&= e^{-\frac{p\epsilon^2}{4}}\\
			&= e^{-\Tilde{c}\epsilon^2 p},
			\end{split}
			\end{equation} where $\Tilde{c}=\frac{1}{4}$. 
			Thus, for any $\Bar{y} \in \mathbb{R}^d$ $\dst P(\big |\|U^T\Bar{y}\|_2^2-\|\Bar{y}\|_2^2 \big | >\epsilon\|\Bar{y}\|_2^2) \leq 2 e^{-\Tilde{c}\epsilon^2p}$. 
			Hence, as stated in \cite{RIPwithJLIEEESP2020}, subspace RIP holds in the transformed space using the proposed projection matrix.
		\end{proof}
	\end{lemma}
	From the proof of Lemma \ref{proofofJLproperty}, the Subspace Restricted Isometry Property (RIP) holds in the transformed space using $U$.
	\subsubsection{Dictionary learning in the transformed space}
	After transformation of data $Y$ into $Z=U^TY$, where 
	$Z\in \mathbb{R}^{p\times N}$, then the problem is to find $D,X$ such that
	\begin{equation}\label{eq:transfeq}
	Z\approx DX + noise, 
	\end{equation} i.e.,
	\begin{equation}\label{eq:transfeq1}
	<D,X>=\argmin_{D,X} \|Z-DX\|_F^2 + \gamma \|X\|_1, 
	\end{equation}
	where $\gamma$ is the regularization parameter controlling the sparsity of the coefficient matrix $X$.
	
	This jointly non-convex problem \eqref{eq:transfeq1} is solved using the alternating minimization method. Initializing $D$ as a random Gaussian matrix ensures the Restricted Isometry property of $D$ \cite{RIP_JL_2020}, \cite{RIPofrandomGaussianDwithJ-L}.
	The second term in \eqref{eq:transfeq1} imposes a sparsity constraint on the coefficient matrix. The sparsity of the coefficient matrix is achieved using Sparse Bayesian Learning, with the assumption of additive Gaussian noise  \cite{SBL_visualtracking2005} in \eqref{eq:transfeq1}. Assuming Gaussian prior on the coefficient matrix, $X$ is optimized using M-SBL \cite{madhuri2023discriminative}. Using M-SBL, irrelevant features are eliminated at the point of convergence, thus making the coefficients sparse. $K-$SVD \cite{ksvd} is used to update dictionary $D$ fixing $X$. This alternating optimization continues when $\|Z-DX\|_F^2 > tolerance$.
	
	In Claim \ref{innerprodclaim}, it is statistically proved that the proposed projection method transforms data so that a structure exists both in the transformed space and the latent feature space of sparse coefficients. This structure in the form of Euclidean distances has been used in several classification (\textit{SVM, k-NN}) methods and clustering (\textit{k-means}) methods. The existence of such Euclidean geometry in both the original and latent feature space, as explained in Claim \ref{innerprodclaim}, ensures that the cluster structure is preserved using JL-lemma for M-SPCA. 
	
	\begin{claim}\label{innerprodclaim}
		The magnitude of the difference between the cosine similarities of projected data and the cosine similarities of corresponding sparse coefficients is bounded. 
		\begin{proof}
			Let $\bar{y_1}$,$\bar{y_2}$ be any two points in the $d-$dimensional space. Let $U^T(\bar{y_1})=\bar{z_1}, U^T(\bar{y_2})= \bar{z_2} \in \mathbb{R}^p$ be their projections where $U\in \mathbb{R}^{d\times p}$ is the transformation matrix. Let $\bar{x_1},\bar{x_2} \in \mathbb{R}^K$ be their sparse coefficients w.r.t dictionary $D \in \mathbb{R}^{p \times K}$ with $K>p$ (since $D$ is an overcomplete shared discriminative dictionary).
			We consider sparse coefficients learnt w.r.t $D\in \mathbb{R}^{p\times K}$, as high dimensional features $\dst(K>p)$. From \cite{dimredusingSR2011}, it is clear that the projection matrix constituted by the eigenvectors corresponding to the largest $p$ eigenvalues of $YLY^T$ transforms data such that 
			$\dst E\bigg[{<\frac{\bar{z_1}}{\|\bar{z_1}\|},\frac{\bar{z_2}}{\|\bar{z_2}\|}> - <\frac{\bar{x_1}}{\|\bar{x_1}\|},\frac{\bar{x_2}}{\|\bar{x_2}\|}>}\bigg]^2$ is bounded. If $X$ is a random variable and $E[X^2]<\infty$ then $Var[X]<\infty$. Thus, by Chebychev inequality, the difference between the cosine similarities of the projected data points and the cosine similarities of their corresponding sparse coefficients is bounded. Thus, the dictionary learned in the M-SPCA embedded space generates similar sparse coefficients of similar data points.
		\end{proof}
	\end{claim}
	
	\subsection{Proposed Classification Rule}\label{sec:propclassifyrule}
	The sparse coefficients obtained from JLSPCADL retain the reconstruction abilities of the $K-$SVD dictionary and the local features due to SPCA. The sparsity pattern is the same for the same class samples because the set of dictionary atoms representing the group is the same for all of them. Hence, these coefficients can be clustered with mean sparse coefficients or medoids of sparse coefficient vectors for each class as cluster centers. The complexity of finding $C$ medoids is $O(C(n_c-1)^2)$, where $n_c$ is the number of sparse coefficients of each class. 
	For offline medoid computation, compute the pairwise distances among the sparse coefficients, compute row-sum or column-sum, return  $\dst \argmin\limits_{x_j=1,2,\ldots,n_c}\sum\limits_{k=1,2,\ldots,n_c, k\neq j} |d(x_j,x_k)|$ as the medoid $\bar{m_c}$ of class $c$.
	Even for online data, medoid is easy to compute when new data of the same class is added, as the old medoid is swapped with the new data point only if the sum of the distances to all the points is minimum for the new data point. Moreover, computing the distances from sparse coefficients with the same sparsity profile is relatively faster. Thus, a small sample size of sparse coefficients reduces the computation complexity of class-wise medoids when new data is added, as demonstrated on the Extended YaleB face image dataset in Table \ref{timelistextyaleb}. 
	The classification rule \eqref{eq:classifyrule} depends on both the reconstruction error and the Euclidean distance between $\bar{x_q}$ and the medoid of sparse coefficients $\bar{m_c}$ of each class. 
	\begin{equation}\label{eq:classifyrule}
	label(\bar{q})=\argmin_{c=1,2,\ldots,C}\{\|\bar{z_q}-D\bar{x_q}\|_2^2 + \tau \|\bar{x_q}-\bar{m_c}\|_2^2\},
	\end{equation}
	
	where $\tau$ is the weightage given to the $l_2-$norm between $\bar{x_q}$ and $\bar{m_c}$.
	The framework for discriminative dictionary learning and image classification is described in Algorithm \ref{JLSPCADL}.
	\begin{algorithm}\caption{JLSPCADL}\label{JLSPCADL}
		\footnotesize
		\begin{algorithmic}[1]
			\State Input:Normalized data $Y$, $\sigma^2, \epsilon$
			\State Output:label($\bar{q}$
			\State Compute $p \geq \frac{12 \log N}{\epsilon^2(1.5-\epsilon)}$ for optimal $\epsilon$.
			\If{$p>d$}
			\State $V$ from \eqref{app:kspca}.
			\State $Z \gets V^T(\Phi(Y)^T\Phi(Y))$, where $\Phi$ is Gaussian kernel.
			\State $D \gets random\_normal\_matrix \in \mathbb{R}^{p \times K}$.
			\State Update $D$ using $K-$SVD s.t. $Z \approx DX$.
			\State $\dst \bar{\alpha}^{(0)} \gets (1,1,\ldots,1)^T$.
			\State $\dst X_1,X_2,\ldots,X_C \gets M-SBL(Y,D, \sigma^2)$ using M-SBL \cite{madhuri2023discriminative}.
			\State $\bar{m_c}\gets KMedoids(X_j),j=1,2,\ldots,C$, i.e. $ {\displaystyle \bar{m_c}=\argmin _{y\in  {X_j}}\sum _{j=1}^{N}|d(y,X_{j})|}$.
			\State Given a query $\bar{q}$, 
			$z_{q} \gets V^T(\Phi(Y)\Phi(\bar{q})$.
			\State Find $\bar{x_{q}}$ of $z_{q}$ such that $\dst \bar{z_q} \approx D{\bar{x_q}}$ using M-SBL \protect \cite{madhuri2023discriminative}.
			\Else
			\State $U$ from \eqref{UequnwoC}.
			\State $\dst Z \gets U^TY$.
			\State Do Steps 7 to 11.
			\State Given a query $\bar{q}$,
			$\dst \bar{z_q} \gets U^T\bar{q}$.
			\State Find ${\bar{x_q}}$ of $\dst \bar{z_q}$ w.r.t $D$ such that $\dst \bar{z_q} \approx D{\bar{x_q}}$ using M-SBL \cite{madhuri2023discriminative}. 
			\EndIf
			\State Classify using \eqref{eq:classifyrule}
		\end{algorithmic}
	\end{algorithm}
	\subsection{Convergence and Complexity analysis }
\begin{figure}
		\centering
		\includegraphics[height=4cm,width=5.5cm]{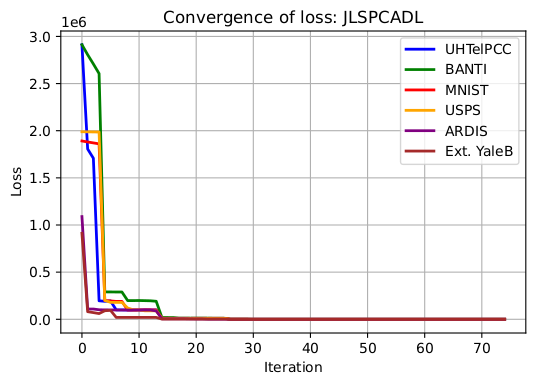}
		\caption{The loss function for dictionary learning, while alternatingly optimizing $D$ (fixing $X$), converges within few iterations.}\label{fig:JLSPDLconvergencegraph}
	\end{figure}

	In JLSPCADL, the last two terms of the objective function are independent of $D, X$. The objective function to derive a single-step solution of the transformation matrix $U$ using SPCA under the orthogonality constraint is convex. The convergence of the loss function using $K$-SVD for dictionary learning (fixing $X$ obtained from sparse coding using M-SBL) is shown in Fig. \ref{fig:JLSPDLconvergencegraph}.
	Table \ref{complexityZ} gives the computational complexity of $Z=U^TY$ when $p\leq d$ and $p> d$. Table \ref{complexitydictlearn} gives the complexities of different methods used to compare with JLSPCADL.
	\begin{table}[!t]
		\footnotesize
		\begin{center}
			\caption{Complexity of Computing Z.
				\label{complexityZ}}
			
			\begin{tabular}{c c }
				\hline
				Method & Complexity\\
				\hline
				SPCA, $p \leq d$  & $\dst O(pN^2+ p^2N+ d^3+ pdN)$\\
				KSPCA,  $p > d$& $\dst O(p^2C+pCN+pN^2+d^3+pdN)$ \\
				\hline
			\end{tabular}
		\end{center}
	\end{table}
	\begin{table}
		\footnotesize
		\begin{center}
			\caption{Complexity of learning Dictionary $D$ using JLSPCADL is low compared to other iterative projection-based methods. Here, $n$ is the number of iterations. }
			\label{complexitydictlearn}
			
			\begin{tabular}{c c}
				\hline
				Method & Complexity \\
				
				\hline
				SDRDL \cite{SDRDL}  & $\dst O(nK^3)+O(ndKN)$\\
				JDDRDL \cite{JDDRDL}  & $\dst O(nK^3)+O(ndKN)$\\
				LC-KSVD \cite{LCKSVD}  & $\dst O(K^3)+O(K^2N+K^2+cKN)$\\
				SEDL\cite{SparseembedDLFace2017PR}&$\dst O(nK^3)+O(ndKN)$\\
				\bf{JLSPCADL}& $\dst O(p^2K)+O(pKN)$ \\
				\hline
			\end{tabular}
		\end{center}
	\end{table}
	The space complexity of \textbf{JLSPCADL} is $\dst O(pd+pN+pK)$ for storing the $U, Z, D$ respectively.
	\section{Experiments and Results}\label{expresults}
	Characteristics of the datasets used for experimentation are given in Table \ref{tab:datainfo}. 
	\begin{table}
		\footnotesize
		\begin{center}\caption{Characteristics of Datasets: No. of classes, cardinalities of majority class, minority class along with imbalance ratios.}\label{tab:datainfo}
			\begin{tabular}{c c c c c}
				
				Data&$\#C$&\textit{$|Maj.class|$} &\textit{$|Minorclass|$}& $Imbal. ratio=|Maj|/|Minor|$
				\\
				\hline
				UHTelPCC \cite{rakeshrtip2r}& $325$ &4392 &20&\bf{219.6}
				\\
				Banti\cite{bantiocr}&$457$ &412 &17&\bf{24.23}
				\\
				MNIST\cite{mnist}&10 &6742 &5421& 1.24\\
				USPS\cite{usps}&10&1194 &542& \bf{2.2}\\
				ARDIS\cite{kusetogullari2020ardis}&10&660 &660& 1\\
				Ext.YaleB\cite{ExtYaleBPAMI2001}& 28&495&445& 1.11\\
				Crop.YaleB\cite{ExtYaleBPAMI2005}&38&56 &45& 1.25\\
				\hline
			\end{tabular}
		\end{center}
	\end{table}
	%
	Ten-fold cross-validation results on Telugu OCR datasets have been obtained on an Intel R Xeon(R) CPU E5-2620 v3 @ 2.40GHz processor with 62.8 GiB RAM. Table \ref{tab:JLSPCADLresults} demonstrates the comparable performance of the proposed JLSPCADL on datasets of different types and sizes. The model is trained with noisy samples added to the minority class, leading to better generalization performance.
	\begin{table}
		\footnotesize
		\caption{Results of JLSPCADL on Telugu OCR and Face recognition datasets are superior compared to dimensionality reduction-based iterative DL methods. SCMLP is an MLP-based method with error back-propagation.}
		\label{tab:JLSPCADLresults}
		\begin{tabular}{p{0.7in} p{0.6in} p{0.4in} p{0.4in} p{0.6in} p{0.5in} p{1in}}
			\hline
			Data &PCA+ LCKSVD \cite{LCKSVD} & PCA+ SEDL \cite{kernelizedsupDL}  & PCA+ SCMLP \cite{SCMLP2019} & JDDRDL \cite{JDDRDL}& SDRDL \cite{SDRDL}&JLSPCADL  $\{F1\}$ \\
			\hline
			UHTelPCC & 74.6  & 90.43  & 99.21&95.89&96.97 & $\bf{99.69 \pm 0.78}$ \{99.43\}\\
			Banti&65.2  & 71.9  & 91&89.65&88.21 &$\bf{91.3\pm0.34}$ \{90.92\} \\
			MNIST& 93.6  & 94.43 &96.5&88.21 & \bf{98.13}&$96.99\pm0.13$ \{96.4\}
			\\
			USPS &91.2  & 96.81  & 96.7&96.45&96.3 &  $\bf{97.2\pm0.56}$  \{96.1\}
			\\
			ARDIS & 90.9 & 93.24 & 94.12&93.65&94.71 & $\bf{95.6\pm0.76}$  \{94.4\}  \\
			Ext. YaleB &90.6  & 95.65 & 97&$66.5\pm0.21$&96.7&  $\bf{99.78\pm0.13}$  \{99.32\}
			\\
			Cropped YaleB& 79.8 & 87& $\bf{96.9}$ &89.65&88.21& $95.46\pm0.89$ \{94.99\}
			\\
			\hline
		\end{tabular}
	\end{table}
	\begin{figure}[!t]
		\vspace*{-2mm}
		\subfloat[\small $p=320$ for UHTelPCC ($d=1024$)]{\includegraphics[width=4.5cm]{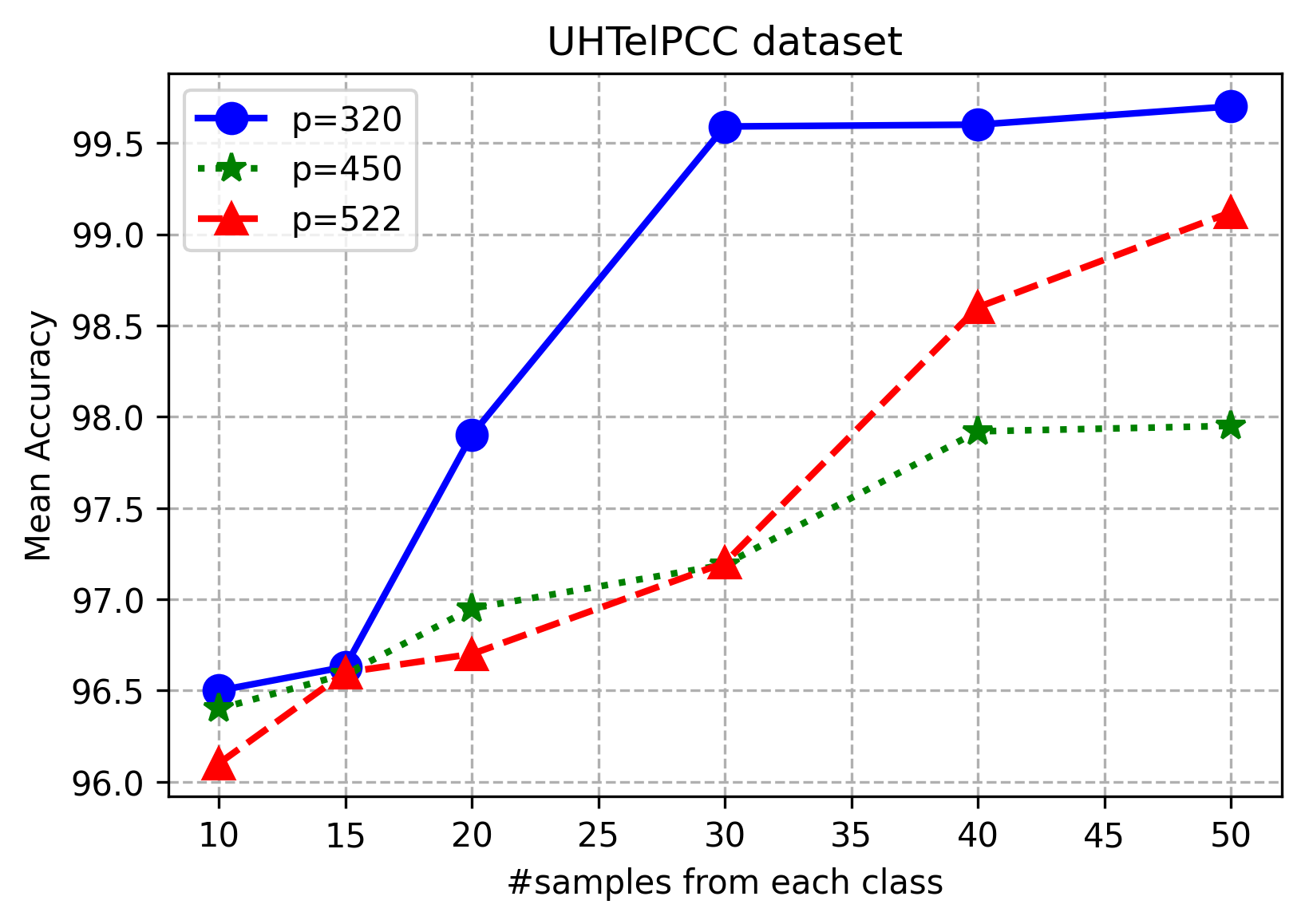}}
		\subfloat[\small $p=319$ for Banti ($d=1024$) ]{\includegraphics[width=4.5cm]{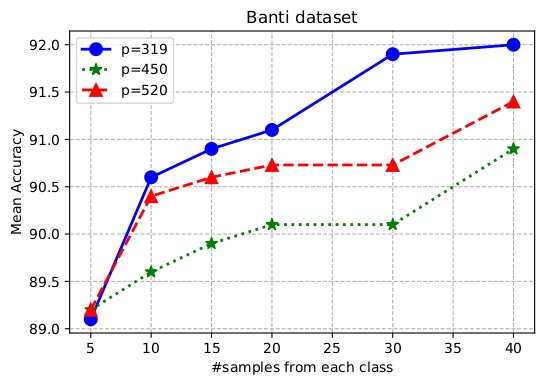}}
		\subfloat[\small Ext. YaleB images with $30\%$ corrupted pixels ]{\includegraphics[height=3cm,width=4.5cm]{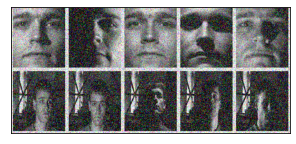}}\\
		\subfloat[\small $p=450$ better for MNIST ($d=784$).]{\includegraphics[width=4.5cm]{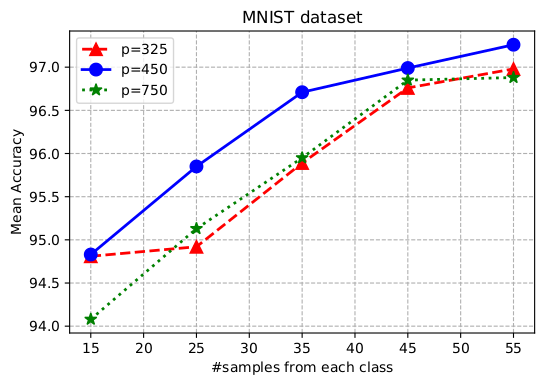}}
		\subfloat[\small $p=264$ better for USPS ($d=256$).]{\includegraphics[width=4.5cm]{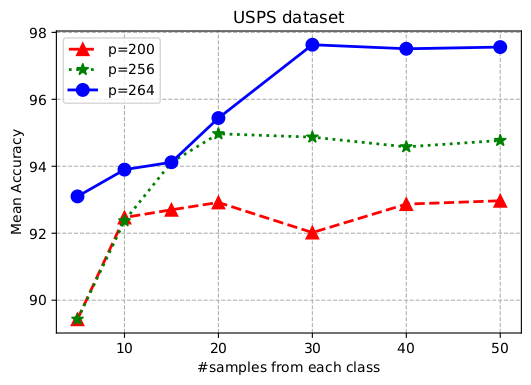}}
		\subfloat[\small $p=424$ better for ARDIS ($d=784$).]{\includegraphics[width=4.5cm]{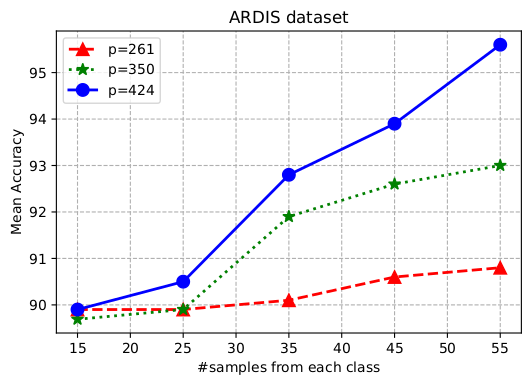}}\\
		\subfloat[\small $p=457$ better on\\ YaleB ($d=192\times 168$)]{\includegraphics[width=4.5cm]{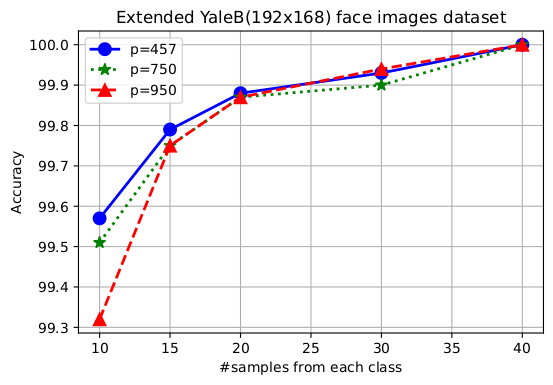}}
		\subfloat[\small $p=457$ better on\\ YaleB ($d=300\times 300$)]{\includegraphics[width=4.5cm]{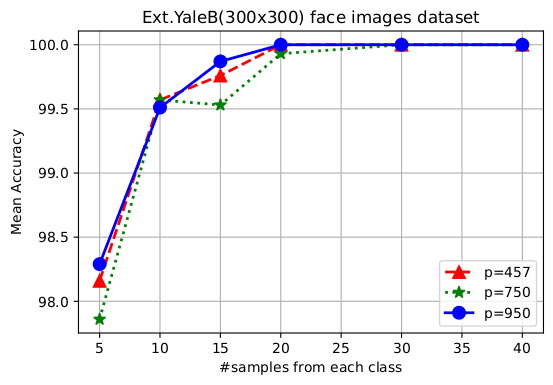}}
		\subfloat[\small $p=365$ better for cropped YaleB ($d=32256$)]{\includegraphics[width=4.5cm]{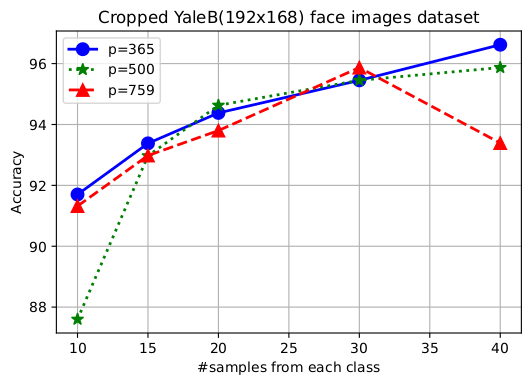}}
		\caption{Random choice of the number of principal components does not work well for discriminative dictionary learning in the transformed space. 
			First row: Classification accuracy when a small sample size from each class is used to form a shared global dictionary using the proposed method.\\ Second row: Classification accuracy for different projection dimensions on handwritten digit datasets.\\ Third row: Classification performance on YaleB is better when the perturbation threshold, $\epsilon$, is low.}
		\label{fig:pvsaccgraphs}
		\vspace{-.5cm}
	\end{figure}
	The number of samples from each class required for training the dictionary under different transformations versus accuracy on Telugu datasets are depicted in  Fig. \ref{fig:pvsaccgraphs}(a), (b). We observe that the misclassification of Telugu printed OCR images is high using PCA+LCKSVD, where a classifier matrix is learned along with the dictionary. 
	\begin{figure}[!t]
		\centering
		\begin{tabular}{lcccc}
			\includegraphics[width=1cm]{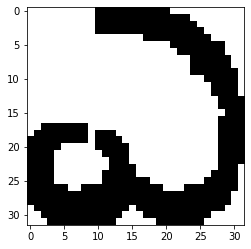}&
			\includegraphics[width=1cm]{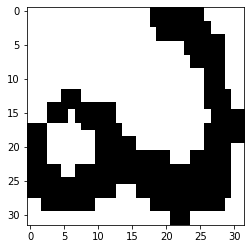}&
			\includegraphics[width=1cm]{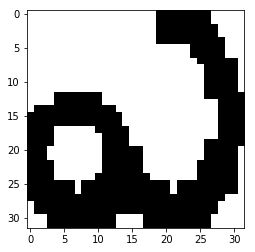}&
			\includegraphics[width=1cm]{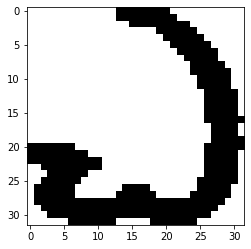}&
			\includegraphics[width=1cm]{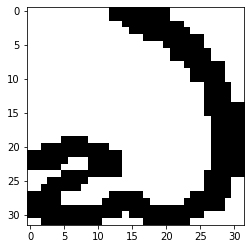}\\
			
			\includegraphics[width=1cm]{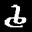}&
			\includegraphics[width=1cm]{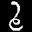}&
			\includegraphics[width=1cm]{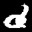}&
			\includegraphics[width=1cm]{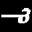}&
			\includegraphics[width=1cm]{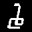}\\
		\end{tabular}
		\caption{The inter-class similarity of UHTelPCC data (first row) and the intra-class variance of Banti data (second row) lead to confusing classes in Telugu OCR data.}\label{confusingpairs_intraclassvar}
	\end{figure} 
	The classification performance of our method on the Telugu dataset UHTelPCC is better, as shown in Fig. \ref{fig:pvsaccgraphs} (a), (b), despite confusing classes (inter-class similarity) as shown in  Fig. \ref{confusingpairs_intraclassvar}. Misclassification of Banti characters could be attributed to different font styles used in the dataset, with high intra-class variability as shown in  Fig. \ref{confusingpairs_intraclassvar}.
	Three-fold cross-validation results on handwritten numerals datasets shown in Fig. \ref{fig:pvsaccgraphs} (d), (e), (f) have been performed on Intel(R) Core(TM) i7-10510U CPU @ 1.80GHz x8 processor with 15.3 GiB RAM. 
	\begin{figure}
		\centering
		\includegraphics[width=1cm]{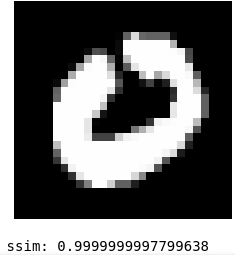}
		\includegraphics[width=1cm]{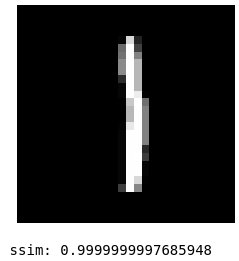}
		\includegraphics[width=1cm]{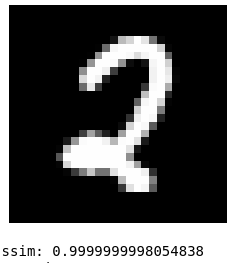}
		\includegraphics[width=1cm]{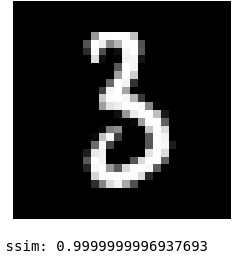}
		\includegraphics[width=1cm]{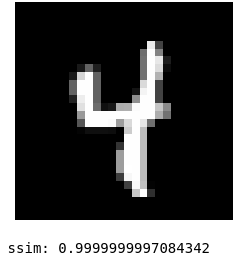}
		\\
		\includegraphics[width=1cm]{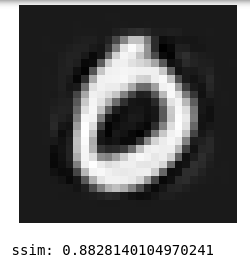}
		\includegraphics[width=1cm]{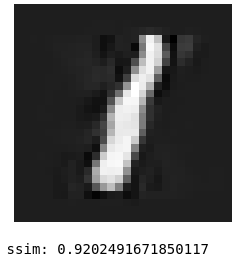}
		\includegraphics[width=1cm]{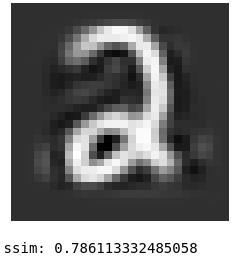}
		\includegraphics[width=1cm]{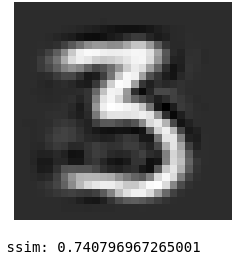}
		\includegraphics[width=1cm]{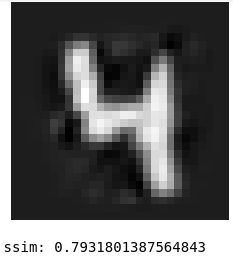}
		\caption{Classification using a shared dictionary requires less time than class-wise residual error-based classification. Sample reconstructed images with SSIs of USPS dataset for $p=256$ (with SPCA), w.r.t. the class-specific dictionaries (First row) and w.r.t the shared discriminative dictionary (second row). }\label{reconstructimages}
	\end{figure}    
	For face recognition datasets, the optimal perturbation threshold interval is $\epsilon \in [0.3,0.4]$, with $p$ decreasing from 457 to 281. However, $p=950,p=750$, and $p=457$ are considered to observe how increasing the projection dimension beyond $457$ and the sample size influences the classification accuracy on the Extended YaleB dataset as shown in  Fig. \ref{fig:pvsaccgraphs} (g),  Fig. \ref{fig:pvsaccgraphs} (h). Though higher dimensions give better accuracy, a much lower $p=457$, for $N=13104$, is considered better due to its consistent performance. Similarly, for the Cropped YaleB dataset, $p=365$ for $ N=1939$ gives a consistent performance, as shown in Fig. \ref{fig:pvsaccgraphs} (i). For Cropped YaleB dataset, we considered $p=365,p=500$, and $p=759$ to observe how the projection dimension and sample size influence the classification accuracy as shown in  Fig. \ref{fig:pvsaccgraphs}(i).
	\section{\vspace{-0.2cm}Discussion}\label{discuss}
	The classification performance of the JLSPCADL method has been compared with PCA+LCKSVD, indicating the improved label consistency of JLSPCADL irrespective of the training sequence and with SEDL to show that a non-orthogonal dictionary learned in the transformed space is better for the classification of the datasets considered here. Unlike other iterative projection methods, which require GPUs, the proposed method classifies comparably well with lean computational facilities. 
	The proposed classification rule can classify corrupted YaleB images with results that are on par with the state of the art. However, the classification performance on synthetic Telugu OCR dataset Banti, with high intra-class variance, is inferior to that of \cite{SCMLP2019}. The classification results on handwritten numerals are inferior to those of CNN-based methods. However, JLSPCADL avoids constructing class-specific dictionaries. With a single global dictionary with shared and class-specific features, the proposed JLSPCADL learns a discriminative dictionary in the optimal feature space. The first row of  Fig. \ref{reconstructimages} are handwritten numerals from the USPS dataset constructed using class-specific dictionaries learned using $K-$SVD (SPCA, when $p=256$ ).
	The second row of  Fig. \ref{reconstructimages} corresponds to the images reconstructed using an over-complete ($K \geq 264$) discriminative dictionary, where the reconstruction quality is not good, but the classification is better than other dimensionality reduction-based DL methods compared here. Training and testing times of JLSPCADL on the Extended YaleB dataset are given in Table \ref{timelistextyaleb}. The decreasing values of training times when new samples are added are due to the time saved in computing the medoids of sparse coefficients of each class. The testing times increase slightly with the increasing dictionary size. 
	\begin{table}
		\footnotesize
		\centering
		\vspace{-4mm}
		\caption{Training time on the Ext. YaleB decreases with increasing sample size as the computation of medoids becomes when samples of the same class are added. Testing times increase with the increase in the shared dictionary size.}
		\label{timelistextyaleb}
		
		\begin{tabular}{c c c}
			\hline
			$\#S_C$& $ Tr. time$(s)& $Testtime/signal$(ms)\\
			\hline
			10 &359.5 & 0.3 \\
			15 & 351.6 & 0.42 \\
			20 & 346.49 & 0.46\\
			30 & 340.1 & 0.47\\
			40 & 329.4 & 0.59 \\
			\hline
		\end{tabular}
	\end{table}
	Compared with other dimensionality reduction-based DL methods as given in Table \ref{tab:noisyalebcomp}, noisy Ext.YaleB.
	Table \ref{tab:noisyalebcomp} compares the classification performance of JLSPCADL on $30\%$ corrupted images of Ext.YaleB dataset. It is observed that JLSPCADL is superior when compared to other dimensionality reduction-based methods and low-rank methods. A comparison of JLSPCADL on noisy YaleB versus JDDRDL's and SDRDL's performance on noiseless data is given in Table \ref{tab:noisyalebcomp}.
	\begin{table}
		\footnotesize
		\centering
		\caption{JLSPCADL performs better than other methods on the noisy Ext. YaleB data.}
		\label{tab:noisyalebcomp}
		\begin{tabular}{c c c c}
			\hline
			Method& $ Tr. time$(s)& Acc.&F1\\
			\hline
			PCA+LCKSVD&320.8& 66.71& 66.71\\
			PCA+SEDL & 320.08 & 78.9&78.9 \\
			SDRDL& 330.9 & 76.7& 76.5 \\
			JDDRDL&329.1 &67 &67 \\
			\bf{JLSPCADL} & 264.8 & \bf{89.9} & \bf{89.9}\\
			\hline
		\end{tabular}
	\end{table}
	\subsection{Parameter sensitivity analysis}
	A detailed parameter sensitivity analysis to assess the change in accuracy levels while changing the parameters in the model.
	\begin{figure*}\vspace{-.7cm}
		\subfloat[]{\includegraphics[height=3cm,width=4cm]{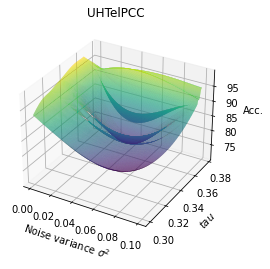}}
		\subfloat[]{\includegraphics[height=3cm,width=4cm]{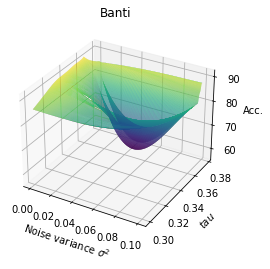}}
		\subfloat[]{\includegraphics[height=3cm,width=4cm]{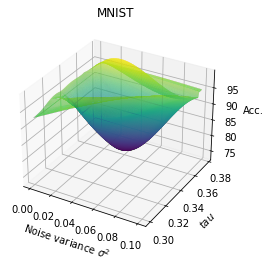}}\\
		\subfloat[]{\includegraphics[height=3.2cm,width=4cm]{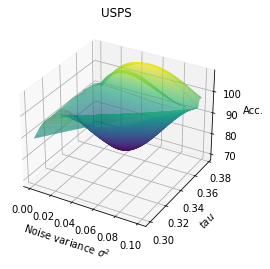}}
		\subfloat[]{\includegraphics[height=3.2cm,width=4cm]{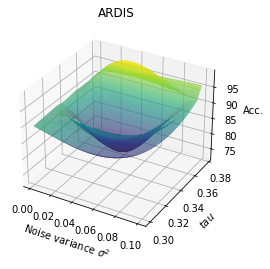}}
		\subfloat[]{\includegraphics[height=3cm,width=4cm]{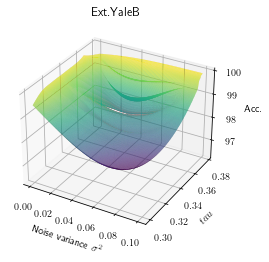}}
		\caption{Effect of noise variance $\sigma^2$ and classification weightage hyperparameter $\tau$ on accuracy. When $\sigma^2$ is between $0.02$ and $0.04$, and $\tau$ is between $0.32$ and $0.38$, the model gives  better classification. For the USPS dataset, $\tau \geq 0.36 $ gives better classification performance.}\label{fig:taunoiseacc}
	\end{figure*}
	Error variance $\sigma^2$ assumed in the Gaussian prior over the coefficient matrix significantly impacts the accuracy levels. A 3d-surface plot of the classification performance of datasets w.r.t the error variance $\sigma^2$ and the classification weightage parameter $\tau$ of \eqref{eq:classifyrule}, is depicted in Fig. \ref{fig:taunoiseacc} using cubic interpolation for unknown function value approximation. The noise variance $\sigma^2$ ranges from $0.001$ to $0.1$ and $\tau$ ranges from $0.3$ to $0.38$. When $\sigma^2$ is between $0.02$ and $0.04$, and $\tau$ is between $0.32$ and $0.38$, the model gives  better classification. For the USPS dataset, $\tau \geq 0.36 $ gives better classification performance. Cubic interpolation gives a smoother and more accurate approximation of the original function than linear interpolation but sometimes overshoots, as in Fig. \ref{fig:taunoiseacc}(d).
	\begin{figure*}\vspace*{-1cm}
		\subfloat[]{\includegraphics[height=3cm,width=4cm]{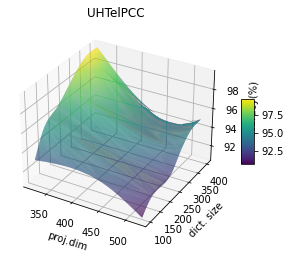}}
		\subfloat[]{\includegraphics[height=3cm,width=4cm]{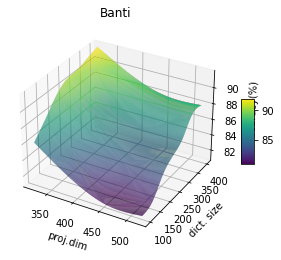}}
		\subfloat[]{\includegraphics[height=3cm,width=4cm]{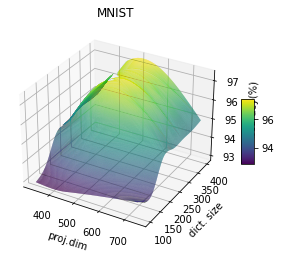}}\\
		\subfloat[]{\includegraphics[height=3cm,width=4cm]{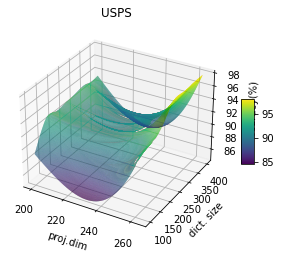}}
		\subfloat[]{\includegraphics[height=3cm,width=4cm]{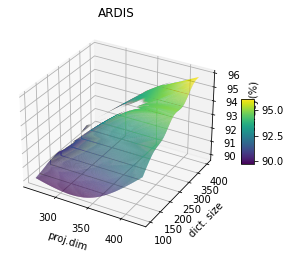}}
		\subfloat[]{\includegraphics[height=3cm,width=4cm]{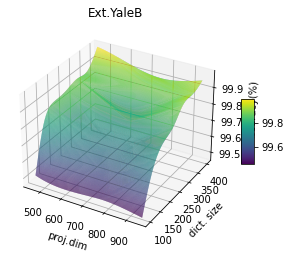}}
		\caption{Effect of projection dimension $p$ and dictionary size $K$ on accuracy. With optimal $p$, even with smaller dictionaries, the model gives better classification. A dictionary with more atoms than $p$ retains the minor details required for classification and thus gives better accuracies.}\label{fig:pKvsacc}
	\end{figure*}
	The effect of the dictionary size $K$ and the projection dimension $p$ such that the dictionary is overcomplete, i.e., $K>>p$, is depicted in Fig. \ref{fig:pKvsacc}. Though the classification accuracy is interpolated to be higher with increasing $K$, it is observed that the optimal perturbation threshold $\epsilon \in [0.3,0.4]$ leading to suitable description length of dictionary atoms ($p$) demonstrates the tradeoff between the dictionary size and better classification of the datasets. 
	\section{\vspace{-1.5mm}Conclusion}\label{conclusions}
	A dimensionality reduction based DL method, JLSPCADL is proposed in this article. The method combines the advantages of Supervised PCA  and the JL-lemma. While the JL-lemma gives the SDL of the dictionary atoms to achieve separable subspaces, the MSPCA gives the JL-prescribed number of components of the transformation matrix with maximised feature-label consistency. The transformed feature space thus preserves the distances between the subspaces and provides for discriminative sparse coefficient extraction.
	Unlike other iterative optimization methods, JLSPCADL obtains the transformation matrix for dimensionality reduction in a single step. It is mathematically proved that the proposed transformation preserves the distances and the angles between the data points. The low dimensional atoms of the global dictionary learned in the transformed space have both global and local features. Sparse coding using this global shared dictionary excludes the irrelevant features and generates discriminative sparse coefficients as features for classification. Due to lower complexity of the method, real-time implementation is possible with small computational facilities. The experimental results on various types of image datasets show that the proposed approach gives better results despite confusing classes, even in the case of highly imbalanced datasets. Under the proposed framework, an approach to learning optimal dictionary size and the atoms could be a research direction. In future work, the Gaussian prior on coefficient vectors could be replaced with a global-local shrinkage prior, leading to correct shrinkage of large signals and noise. 
	\appendix
	\addcontentsline{toc}{section}{Appendices}
	\subsection{Kernel SPCA}
	When $p>d,$ Kernel SPCA \cite{supPCAPR2011}  has been used. The transformation matrix $U$ is expressed as a linear combination of the projected data points. Let $\Phi:$ $\bar{y}$ $\to$ $feature\_space$, then $U=\Phi(Y)V$. The objective is to find $U$, which maximizes the dependence between $\Phi(Y)$ and the label matrix $H$. After rearranging as in \eqref{UequnwoC},
	\begin{equation}\label{app:kspca}
	\begin{split}
	\Hat{U} = \argmax_{U} \{tr(U^T \Phi(Y)H^TH \Phi(Y)^TU)\}\\
	=\argmax_{V} \{ tr(V^T\Phi(Y)^T\Phi(Y)H^T\Phi(Y)^T\Phi(Y)V)\}\\
	=\argmax_{V}\{ tr(V^TK_1H^THK_1V)\}, 
	\end{split} 
	\end{equation} 
	sub. to $U^TU=I$ $\implies$ $\dst V^TK_1V =I$  where $K_1=\Phi(Y)^T\Phi(Y)$. $\Phi(Y)V$ is a semi-orthogonal matrix and hence is an isometry. The generalized eigenvector problem is solved to get $V=$ $p$ eigenvectors corresponding to the top $p$ eigenvalues of $\dst [K_1H^THK_1, K_1]$. The transformed space is given by $Z= (\Phi(Y)V)^T\Phi(Y)$ i.e. $\dst Z=V^TK_1$ whose complexity is $\dst O(Np^2+pN^2)$. $\dst Z_{test}=V^T\Phi(Y)^T\Phi(y_{test})$.
	In the dual formulation problem, find  SVD of $\Phi(Y)H^T= U\Sigma V^T$ and the left-singular vectors of $\Phi(Y)H$ are eigenvectors of $\Phi(Y)HH^T(\Phi(Y))^T$. In SVD, the first $p$ columns in $U$ corresponding to the first $p$ non-zero singular values are in a one-to-one correspondence with the first $p$ rows of $V^T$. Thus, we can reduce $U,\Sigma, V^T$ to $U^1, \Sigma^1, (V^1)^T$ to get the same result. $U^1= \Phi(Y)H V^1 (\Sigma^1)^{-1}$ and $Z= (U^1)^T\Phi(Y)\implies$ $ Z= (\Sigma^1)^{-1}(V^1)^TH^T\Phi(Y)^T\Phi(Y)$. So, the complexity is $O(p^2C+pCN+pN^2)$ where $C$ is the number of classes. 
	\subsection*{Acknowledgements}
	\vspace*{-1mm}
	The authors would like to thank Dr. Sumohana S. Chanappayya for the relevant suggestions for improving the article.
	\bibliography{ISproposed.bib}
	\bibliographystyle{elsarticle-harv}     
\end{document}